\documentclass[conference]{IEEEtran}
\IEEEoverridecommandlockouts
\usepackage{cite}
\usepackage{amsmath,amssymb,amsfonts}
\usepackage{graphicx}
\usepackage{textcomp}
\usepackage{xcolor}
\usepackage{url}
\usepackage{array}
\usepackage{multicol}
\usepackage{hyperref}
\usepackage{algorithm}
\usepackage{algorithmic}
  \definecolor{score1}{RGB}{164,57,49}
 \definecolor{score2}{RGB}{240,146,68}
 \definecolor{score3}{RGB}{23,162,184}
 \definecolor{content}{RGB}{186,53,49}
 \definecolor{metadata}{RGB}{243,121,62}
 \definecolor{references}{RGB}{57,166,57}
 \definecolor{output}{RGB}{59,94,110}
 \definecolor{article}{RGB}{255,153,104}
 \definecolor{law}{RGB}{209,145,199}
 \definecolor{decree}{RGB}{87,149,129}
 \definecolor{code}{RGB}{250,102,103}
\def\BibTeX{{\rm B\kern-.05em{\sc i\kern-.025em b}\kern-.08em
    T\kern-.1667em\lower.7ex\hbox{E}\kern-.125emX}}
\begin{document}

\title{An Senegalese Legal Texts Structuration Using LLM-augmented Knowledge Graph
}

\author{
	\IEEEauthorblockN{Oumar Kane\IEEEauthorrefmark{1}, Mouhamad M. Allaya\IEEEauthorrefmark{2}\IEEEauthorrefmark{4}, Dame Samb\IEEEauthorrefmark{2}\IEEEauthorrefmark{5}, Mamadou Bousso\IEEEauthorrefmark{2}\IEEEauthorrefmark{3}}
\IEEEauthorblockA{\IEEEauthorrefmark{1}\textit{Sciences and Technologies  | Economic and Social Sciences}, 
\textit{Iba Der Thiam University}, 
Thies, Senegal \\
Email: oumar.kane@univ-thies.sn}
\IEEEauthorblockA{\IEEEauthorrefmark{2}\textit{Economic and Social Sciences}, 
\textit{Iba Der Thiam University}, 
Thies, Senegal}
\IEEEauthorblockA{\IEEEauthorrefmark{4}Email: mouhamad.allaya@univ-thies.sn}
\IEEEauthorblockA{\IEEEauthorrefmark{5}Email: dsamb@univ-thies.sn}\IEEEauthorblockA{\IEEEauthorrefmark{3}Email: mbousso@univ-thies.sn}}

\maketitle

\begin{abstract}
This study examines the application of artificial intelligence (AI) and large language models (LLM) to improve access to legal texts in Senegal's judicial system. The emphasis is on the difficulties of extracting and organizing legal documents, highlighting the need for better access to judicial information. The research successfully extracted 7,967 articles from various legal documents, particularly focusing on the Land and Public Domain Code. A detailed graph database was developed, which contains 2,872 nodes and 10,774 relationships, aiding in the visualization of interconnections within legal texts. In addition, advanced triple extraction techniques were utilized for knowledge, demonstrating the effectiveness of models such as GPT-4o, GPT-4, and Mistral-Large in identifying relationships and relevant metadata. Through these technologies, the aim is to create a solid framework that allows Senegalese citizens and legal professionals to more effectively understand their rights and responsibilities.

\end{abstract}


\section{Introduction}

Artificial intelligence (AI) is a transformative technology that raises significant ethical considerations regarding its use. Initiatives like Microsoft's "AI for Humanitarian Action" and Google's "AI for Social Good" focus on enhancing jurisprudence and human rights \cite{norvig_intelligence_2021}. Moreover, the Center for Social Good Data Science at the University of Chicago applies AI to improve criminal justice systems.

Large Language Models (LLMs) exemplify advanced AI systems that can understand and generate human-like text by grasping syntax and semantics \cite{bouchard_building_2024}. The foundation of leading models such as GPT-4, Claude, and LLaMA stems from transformer-based architectures introduced in the influential paper "Attention Is All You Need" \cite{vaswani_attention_2017, zhao_survey_2023}.

In Senegal, AI has the potential to streamline the access and citation of legal texts effectively. A significant challenge in this endeavor lies in the complexity of navigating diverse regulatory documents that are often interconnected or contradictory \cite{senegal_recueil_2019}. To achieve accurate retrieval and analysis of legal texts, innovative methods must be developed, leveraging LLMs to present information clearly to both citizens and legal practitioners \cite{soh_building_2021}.
 
Numerous studies have explored the organization of legal documents globally. In 2022, a knowledge graph was developed using the Indian Legal Domain Corpus to aid legal scholars and professionals, yielding favorable outcomes from entity extraction and relationship assessment \cite{jain_constructing_2022}. Additionally, in 2023, a robust knowledge graph was generated through a three-step process involving data collection, entity extraction, and heterogeneous graph representation, enabling advanced applications for legal professionals \cite{vuong_constructing_2023}.

This article posits that Large Language Models (LLMs) can elucidate existing connections in legal texts by structuring them into a network of legal entities. An initial rule-based process extracts legal articles and metadata per Senegal's legislative guidelines \cite{senegalese_national_assembly_guide_2011}, which are then stored in a relational database. Moreover, a graph database focused on land rights is systematically constructed using Neo4j \cite{dong_knowledge_2021}. Our research primarily aims to develop a knowledge graph based on LLMs, adhering to the LLM-augmented Knowledge Graph principle (also known as LLM-augmented KG) \cite{ibrahim_survey_2024}. This involves constructing the knowledge graph through various techniques including coreference resolution, named entity recognition, and entity relationship identification \cite{caciularu_cdlm_2021, joshi_spanbert_2020, joshi_bert_2019, peters_deep_2018, onoe_modeling_2021, li_efficient_2020}. Furthermore, relationships within documents can be systematically extracted \cite{ma_dreeam_2023}. In our approach, the LLM-augmented KG is formed by generating knowledge triples through advanced prompting techniques \cite{bouchard_building_2024}, with evaluation conducted on verified samples from the Neo4j graph database.

Overall, our goal is to develop a comprehensive framework that enhances the use of LLMs in addressing legal inquiries from Senegalese citizens and experts regarding their rights and obligations.

The paper is organized as follows: an overview of Senegalese legal text structure, a detailed methodology presentation, an extensive analysis of results, a discussion of significant findings, and a concluding summary of key outcomes.

\section{Structure of Legal Texts in General}\label{document_structure}

This section delineates the architecture of legal documents, encompassing extensive legal codes relevant to specific domains of law. It will highlight the arrangement of legal texts related to land and public domain, recognized for its intricacy. Additionally, it will consider other significant legal codes.

\subsection{General Principle}

The Senegalese legal code includes laws and decrees, each with primary and supplementary articles. Legislative articles, approved by Parliament, specify legal issues, including rights and obligations. Conversely, decree articles function as administrative tools implemented by executive bodies to uphold laws. These articles are initially written in French.

This research examines the structure of these articles, which may be formatted as tables, paragraphs, or bullet points. Each article is timestamped according to the enactment date of its corresponding law or decree and may feature a title. Amendments or replacements to an article are noted at its outset, citing the current law or decree responsible for the change.

Additionally, articles frequently cite other legal provisions, establishing an intricate web of interconnected laws, decrees, and legal codes.

\subsection{Subdivisions}\label{subdivisions}

\begin{figure}[h]
    \centering
    \includegraphics[width=\linewidth]{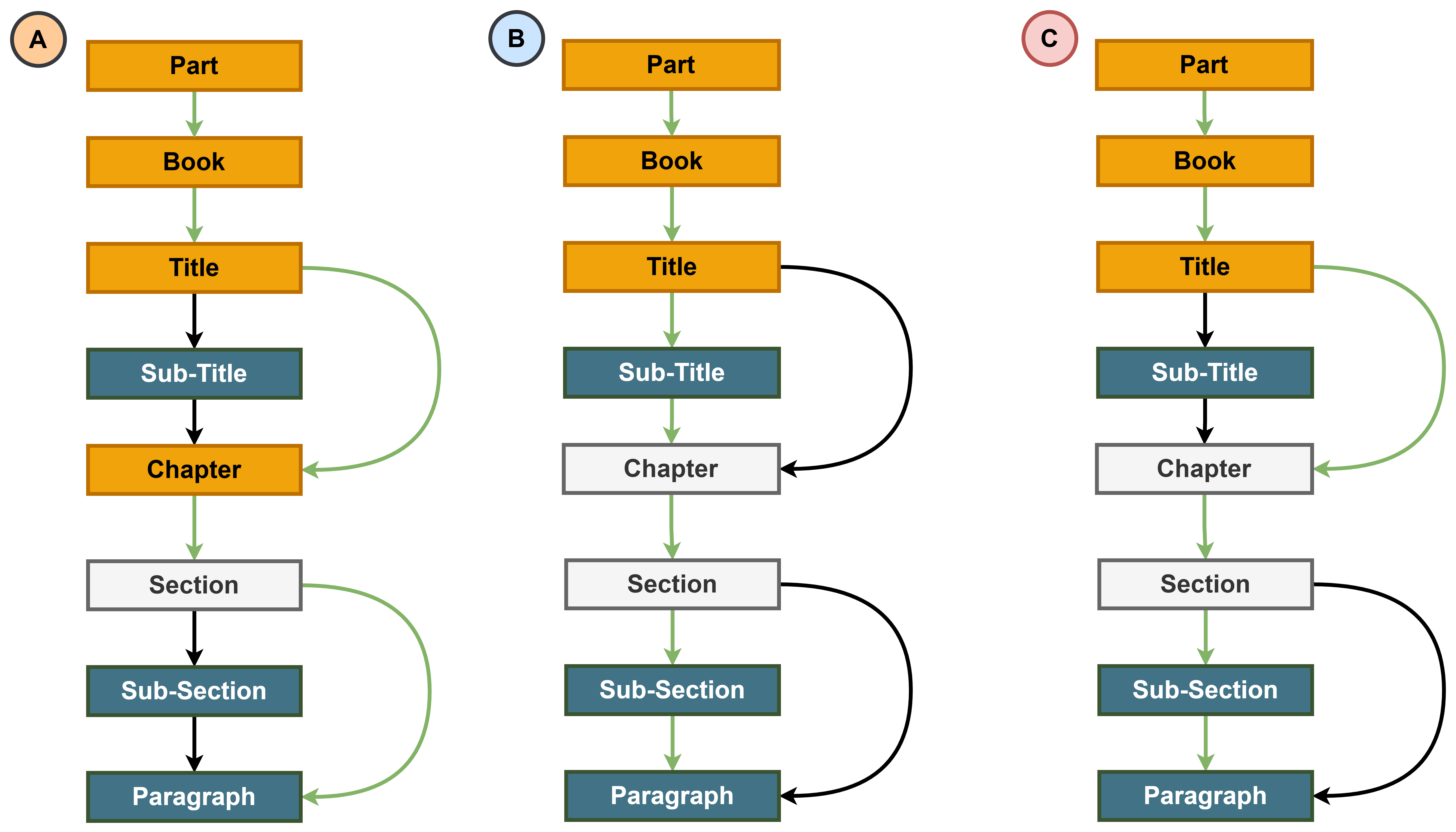}
    \caption{\textbf{Hierarchies of Subdivisions}. The green arrows illustrate the primary pathways of the architectural design. Architecture B pertains to the land and public domain code, along with the public markets code, whereas the majority of legal codes conform to architectures A and C.}
    \label{fig:subdivisions}
\end{figure}

Legislation and official documents consist of various components, including parts, books, titles, and paragraphs. 
These subcategories follow three hierarchical structures, as depicted in Fig. \ref{fig:subdivisions}, with optional subdivisions indicated by orange blocks and dependencies shown in blue blocks. Most codes align with Hierarchies A and C, while the Land and Public Domain and Public Market codes conform to Hierarchy B.

In the Land and Public Domain compilation, rental price regulations are organized into two sections, each containing title subdivisions. The first section specifies localities within a region for each title, while the second section categorizes rent using alphanumeric or numerical formats. Additionally, a cross-tabulation illustrates rent prices across categories and regions.

\subsection{Numbering}

The provisions are enumerated in a systematic fashion within Laws or Decrees, which are likewise assigned numerical identifiers for the purposes of facilitating identification, organization, and longitudinal tracking of amendments. In jurisdictions such as Senegal and France, Laws and Decrees are allocated distinctive numbers predicated on the year of enactment and their sequential position within that specific year, exemplified by Decree 2020-567.

Multiplicative adverbs such as 'bis', 'ter', and 'quarter' may be appended to articles to signify repetitions, whereas prefixes such as R. or L. serve to indicate whether the provision is legislative or regulatory. For instance, "Article L. 1" denotes the primary article within the legislative corpus, while "article 5 bis" reflects a modification or augmentation of the original article designated as number 5.

\section{Methodology}

\subsection{Legal Texts Extraction Algorithm}\label{extract_articles}

An algorithm was developed to extract articles and metadata from legal documents. The extraction procedure for a file named \(doc\_name\) is outlined in Algorithm \ref{extract_document}. Preliminary processing categorized documents into thematic directories and subcategories for laws. Rent-related sections were identified using the boolean variable \(is\_rent = True\). All DOCX files adhere to Senegalese drafting guidelines \cite{senegalese_national_assembly_guide_2011}.

The algorithm produces dictionaries that represent articles. Metadata \(domain\) and \(law\_num\) are derived from \(dir\_path\), while \(name\), \(number\), and \(signature\_date\) are sourced from \(doc\_name\). Files that are unparsable and lack "code" in their name are classified as containing declarative articles.

Two variables, \(content\) and \(content\_r\), store general and rent-specific content. The \(attribs\) dictionary captures metadata for each article, while \(subdiv\_data\) manages hierarchical subdivisions. Conditional flags regulate the processing flow.

The primary function involves iterating through DOCX elements to identify subdivision and article markers. Lower hierarchy subdivisions are removed accordingly. Specialized subdivisions such as housing type and rent category facilitate the identification of rental articles. Completed articles update attributes and are appended to the list of articles.

Additionally, certain attributes were extracted but not explicitly included in Algorithm \ref{extract_document}. These attributes pertain to the categorization of documents as 'Regulatory' or 'Legislative' predicated on \(art\_num\), in conjunction with the boolean attribute \(declaration\) for material pertinent to declarations.



\begin{figure}[h]
    \centering
    \includegraphics[width=1\linewidth]{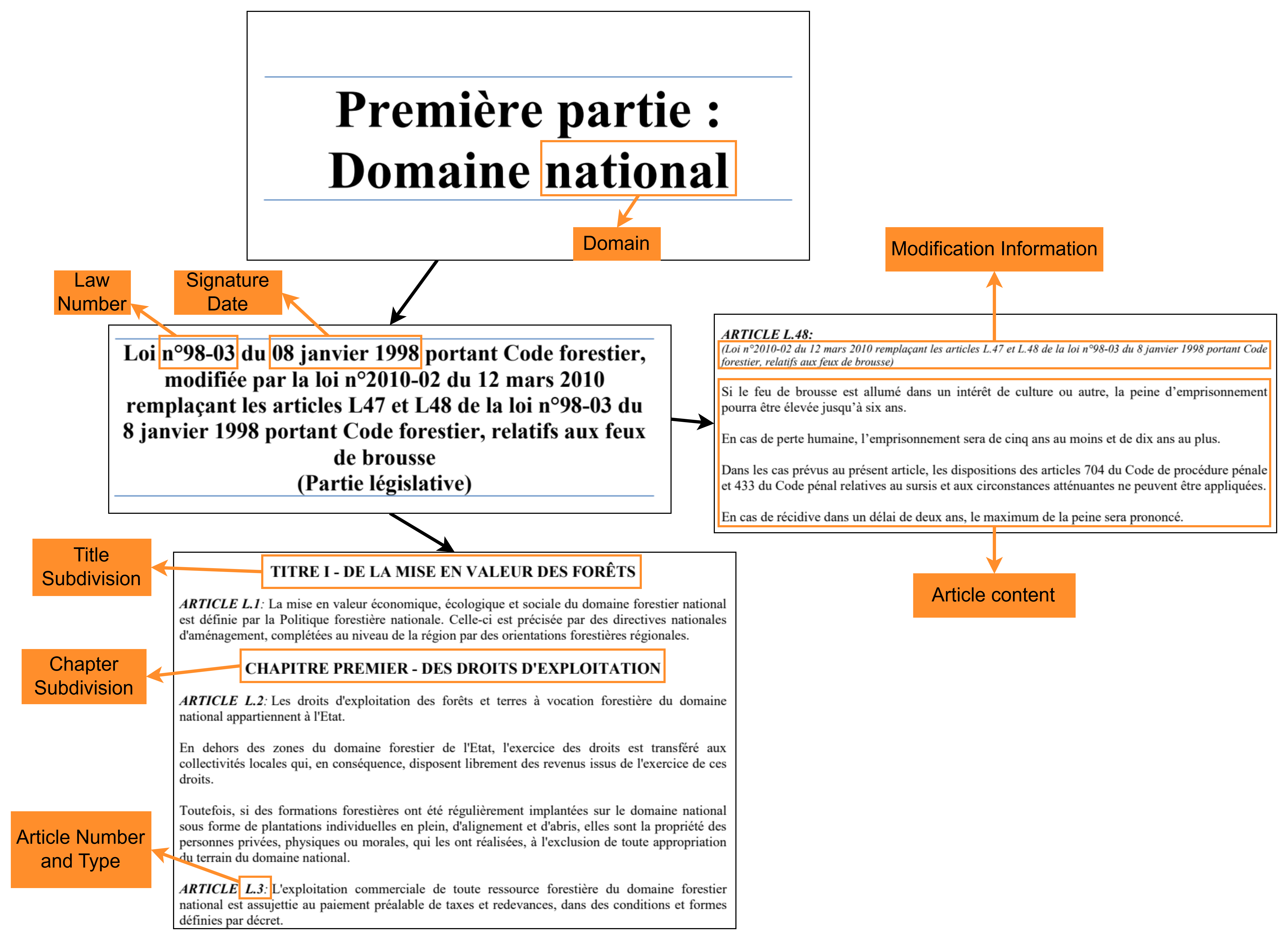}
    \caption{\textbf{Metadata Identification}. This figure demonstrates metadata derived from the land and public domain collection. Direct properties are extracted from the text. Other metadata, including subsections and article titles, are omitted in this instance but are integral to the extraction process.}
    \label{fig:metadata}
\end{figure}

\begin{algorithm}[h]
\scriptsize
\caption{Extract Document Articles}
\label{extract_document}
\begin{algorithmic}[1]
\REQUIRE Directory path $dir\_path$, Document name $doc\_name$, Boolean $is\_rent$
\ENSURE List of article dictionaries $arts$

\STATE Parse $dir\_path$ to extract $domain$, $law\_num$
\STATE Segregate and refine $doc\_name$ to get $name$, $number$, $signature\_date$
\STATE Load document content from $dir\_path + doc\_name$
\STATE Initialize list of articles $arts \gets []$
\STATE Initialize contents $content, content\_r \gets ''$
\STATE Initialize attributes $attribs, subdiv\_data \gets \{\}$
\STATE Initialize flags $in\_r\_subs, in\_art, in\_subdiv, begin\_r \gets$ \texttt{False}

\FORALL{$\epsilon$ in document content}
    \IF{$\epsilon$ denotes a subdivision}
        \STATE Update $subdiv\_data$, 
        \STATE Reset $in\_r\_subs$
        \STATE Set $in\_subdiv \gets$ \texttt{True}
    \ELSIF{$in\_art$ and $in\_subdiv$ and $is\_rent$}
        \IF{$begin\_r$}
            \STATE Add $content\_r$ to $attribs$
            \STATE Append $attribs$ to $arts$
            \STATE Reset $begin\_r$, $content\_r$
        \ENDIF
        \IF{$\epsilon$ denotes rent-specific subdivision}
            \STATE Update $subdiv\_data$
            \STATE Set $in\_r\_subs \gets$ \texttt{True}
        \ENDIF
    \ELSIF{$\epsilon$ marks article start}
        \STATE Set $in\_art \gets$ \texttt{True}
        \STATE Update $attribs$ with $content$
        \STATE Extract $art\_num$, $heading$, $content'$ from $\epsilon$
        \STATE Detect $multiplicative$ in $heading$
        \STATE Reset $in\_subdiv$, $content$
        \STATE Append $content'$ to $content$
        \STATE Append $attribs$ to $arts$
        \STATE Update $attribs$ with $multiplicative$, $art\_num$, and other metadata
        \IF{$is\_rent$}
            \STATE Reset $subdiv\_data$
        \ELSE
            \STATE Update $attribs$ with $subdiv\_data$
        \ENDIF
    \ELSIF{$\epsilon$ is non-empty}
        \STATE Append $\epsilon$ to $content$ or $content\_r$ depending on $in\_r\_subs$, set $begin\_r \gets$ \texttt{True} if appended to $content\_r$
    \ENDIF
\ENDFOR

\IF{$content \neq ''$ \OR $content\_r \neq ''$}
    \STATE Add remaining content to $attribs$, append to $arts$
\ENDIF

\RETURN $arts$
\end{algorithmic}
\end{algorithm}

\subsection{Land and Public Domain Graph Database Development}

To establish a graph database for land and public domain law, primary node labels were identified from the legal document's table of contents. The labels included: \textit{Domain}, \textit{Law}, \textit{Decree}, \textit{Article}, \textit{Official Journal}, \textit{Ministerial Order}, \textit{Declaration}, \textit{Uniform Act}, and \textit{Legal Code}.

Moreover, relationships among these nodes were delineated, such as \textit{publish} (an official journal publishing multiple laws, decrees, and ministerial orders), \textit{possess} (a domain possessing various nodes), \textit{is associated} (a law associated with multiple decrees), and other interactions like \textit{modify} or \textit{repeal} (laws interacting with other laws or articles), as well as \textit{frame} or \textit{execute} (decrees framing or executing laws), and \textit{based on} (articles referencing nodes).

Additionally, the analysis of the table of contents revealed that a \textit{Law}, \textit{Decree}, or \textit{Ministerial Order} could be \textit{signed} by multiple individuals, prompting the incorporation of \textit{Person} as a new node. It was also noted that laws, decrees, ministerial orders, or declarations could possess numerous articles, leading to the creation of possession relationships based on articles extracted as detailed in Subsection \ref{extract_articles}.

To enhance the reliability of the nodes, attributes or metadata were introduced. For instance, attributes such as \textit{title} and \textit{name} were appended to the \textit{Person} node (e.g., title as President and name "Abdoulaye Wade"), while the \textit{object} attribute was affixed to the \textit{Law}, \textit{Decree}, and \textit{Ministerial Order}, \textit{signature date} to the \textit{Official Journal}, and \textit{pages} to the \textit{publish} relationship.

\subsection{Knowledge Triples Extraction with LLMs}

\begin{figure}[h]
    \centering
    \includegraphics[width=1\linewidth]{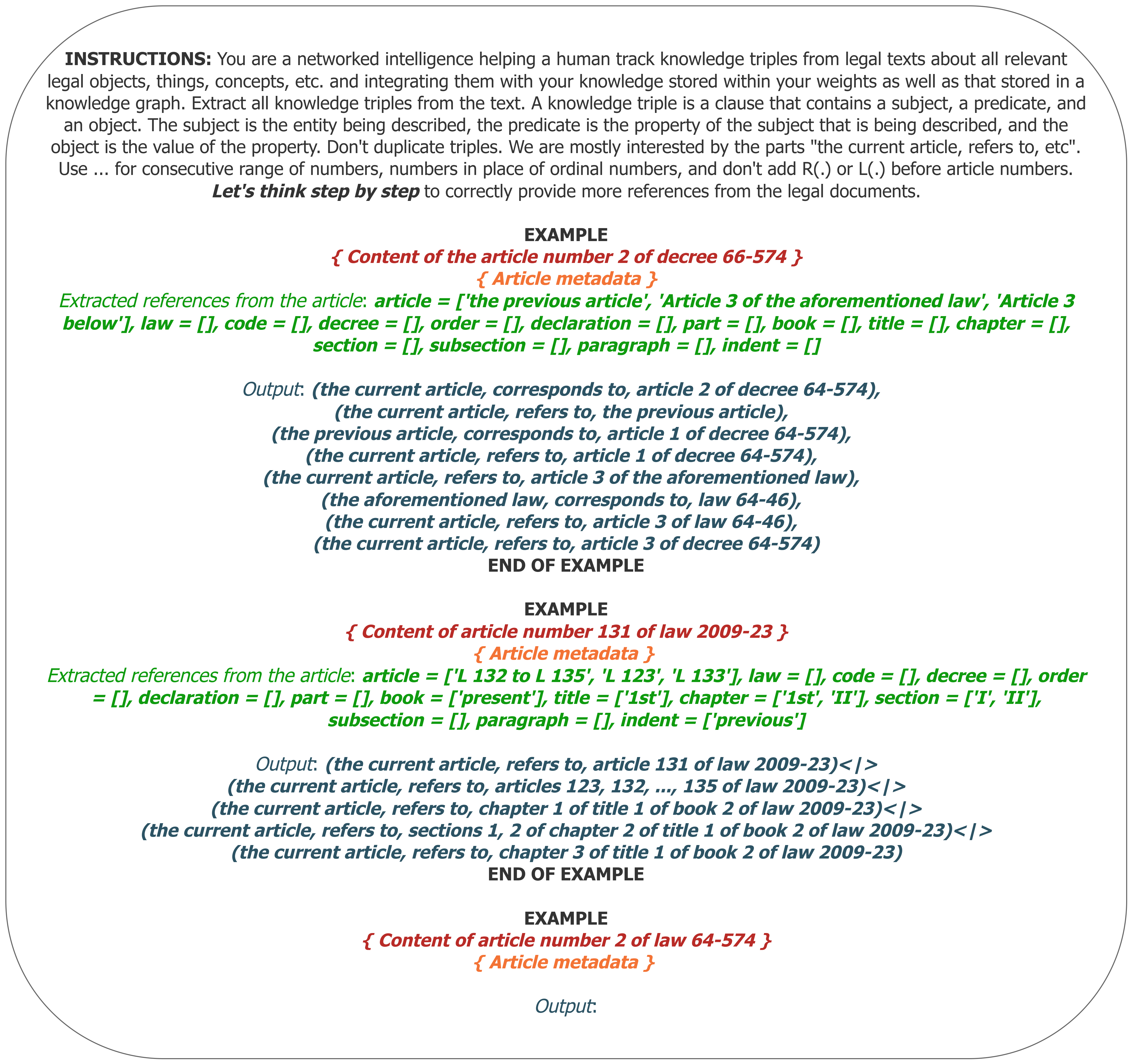}
    \caption{\textbf{Instance of a prompt with two examples for extracting knowledge triples from legal articles.} Each example includes the article content (in \textbf{\textcolor{content}{red}}), article metadata (in \textbf{\textcolor{metadata}{orange}}), and references extracted using LLM parsers (in \textbf{\textcolor{references}{green}}). The output consists of various triples (in \textbf{\textcolor{output}{blue}}). We provide only the Output section for the current article, allowing the LLM to predict the corresponding triples based on the examples. Each triple typically has 'the current article' as the subject, with predicates like 'refers to' or 'corresponds to,' and the object as the legal entity. The structure specifies that "article" or "articles" is followed by their respective number(s), separated by commas. It includes 'of law' plus the law number or 'of decree' plus the decree number. Consecutive article numbers are abbreviated with an ellipsis.}
    \label{fig:triple_prompt}
\end{figure}

Emergent abilities in large language models (LLMs) denote unexpected skill development correlated with model size. These skills, such as arithmetic and summarization, are not directly taught during training. Rather, they appear spontaneously with increased model dimensions. Such capabilities can be applied to extract knowledge triples from legal texts. For instance, we can direct the LLM's output through prompts \cite{bouchard_building_2024}. We furnish the LLM with contextual instructions to facilitate the generation of human-like text. In this study, we employ a Few-Shot Chain of Thought (Few-Shot-CoT) approach \cite{kojima_large_2023}, integrating Few-Shot and Chain-of-Thought \cite{wei_chain--thought_2023} prompting methodologies. Specifically, we present examples to illustrate reasoning for knowledge triple extraction and provide instructions to clarify the objective. We incorporate a "Let's think step by step" \cite{kojima_large_2023} directive in the original prompt to augment its effectiveness. An example of a prompt utilized with LLMs is depicted in Fig. \ref{fig:triple_prompt}.

The highlighted references enhance context and support knowledge triple extraction. Language models structure these references. An output parser delineates the model's expected output. For instance, citations for legal entities guide the model in extracting triples from the article's content.

Furthermore, we focused on utilizing large open models like those from OpenAI or Mistral for reference and knowledge triple extraction.

\section{Experiments and Results}

\subsection{Extracted Legal Documents}

The extraction process yielded a significant volume of articles from twenty legal documents. A total of 7,967 articles were compiled, encompassing various laws, decrees, ministerial orders, and declarations. The majority of articles were derived from the Land and Public Domain Code, as presented in Table \ref{tab:legal_documents}. This table enumerates the articles extracted from each legal document alongside the counts for other legal entities. Additionally, it specifies the years of the latest revisions for these codes. These documents were sourced from various platforms, particularly the Senegalese government websites.

\begin{table}[h]
    \centering
    \caption{Overview of Legal Documents}
    \label{tab:legal_documents}
    \resizebox{\columnwidth}{!}{%
    \begin{tabular}{ccccccc}
    \hline
         \textbf{Document Title}&  \textbf{Last Up.}&  \textbf{Art.}&  \textbf{Laws}&  \textbf{Decrees}&  \textbf{Orders}& \textbf{Dec.}\\
    \hline
         Land and Pub. D. Collection&  2013&  2039&  26&  32&  1& 1\\
    \hline
         Public Procurement Code&  2022&  154&  0&  1&  0& 0\\
    \hline
         Criminal Procedures Code&  N/A&  780&  14&  0&  0& 0\\
    \hline
         Criminal Code&  N/A&  557&  21&  0&  0& 0\\
    \hline
         Local Authorities Code&  1996&  370&  3&  0&  0& 0\\
    \hline
         Family Code&  1999&  851&  3&  1&  0& 0\\
    \hline
         Construction Code&  2010&  379&  1&  1&  0& 0\\
    \hline
         Sanitation Code&  2009&  108&  1&  0&  0& 0\\
    \hline
         Water Code&  1981&  86&  1&  0&  0& 0\\
    \hline
         Hygiene Code&  1983&  81&  1&  0&  0& 0\\
    \hline
         Aviation Code&  2015&  240&  1&  0&  0& 0\\
    \hline
         Investment Code&  2004&  47&  1&  1&  0& 0\\
    \hline
         Mining Code&  2016&  240&  1&  1&  0& 0\\
    \hline
         Civil Procedure Code&  N/A&  920&  0&  4&  1& 0\\
    \hline
         Customs Code&  2014&  427&  1&  0&  0& 0\\
    \hline
         Environment Code&  2013&  237&  1&  0&  0& 0\\
    \hline
       Postal Code&  2006&  48&  1&  0&  0& 0\\
    \hline
       Telecommunications Code&  2001&  77&  1&  0&  0& 0\\
    \hline
       Nationality Code&  1989&  37&  5&  0&  0& 0\\
    \hline
       Labor Code&  1997&  289&  1&  0&  0& 0\\
    \hline
    \end{tabular}}
\end{table}

\subsection{Graph Database}

\begin{figure}[h]
    \centering
    \includegraphics[width=1\linewidth]{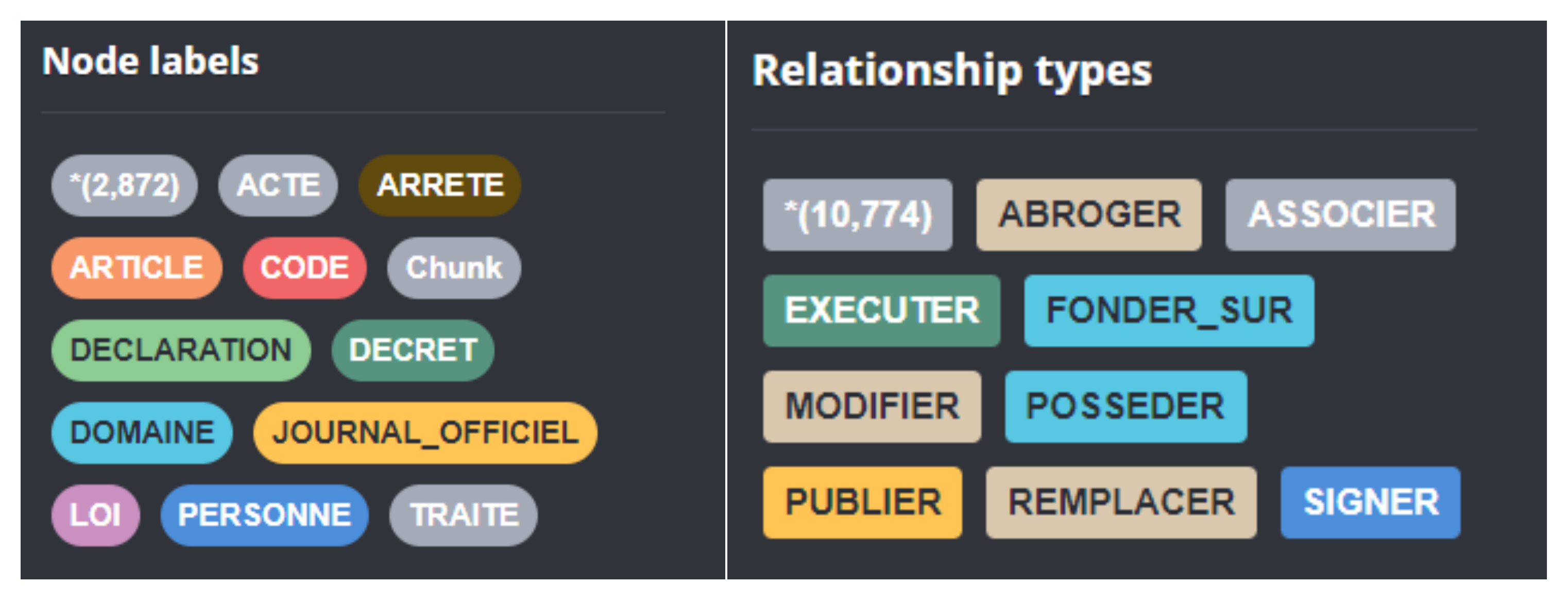}
    \caption{\textbf{Neo4j Database Summary}: The node and relationship types are written in French, as the legal documents were authored in that language. The database contains 2,872 nodes and 10,774 relationship types.}
    \label{fig:neo4j}
\end{figure}

By carefully identifying nodes and relationships in the Land and Public Domain Code, we established a comprehensive graph database comprising 2,872 nodes and 10,774 relationships. Figure \ref{fig:neo4j} illustrates selected information from the Neo4j panel. The nomenclature of nodes and relationships is presented in French to enhance comprehension for Senegalese legal professionals.

\begin{figure}[h]
    \centering
    \includegraphics[width=1\linewidth]{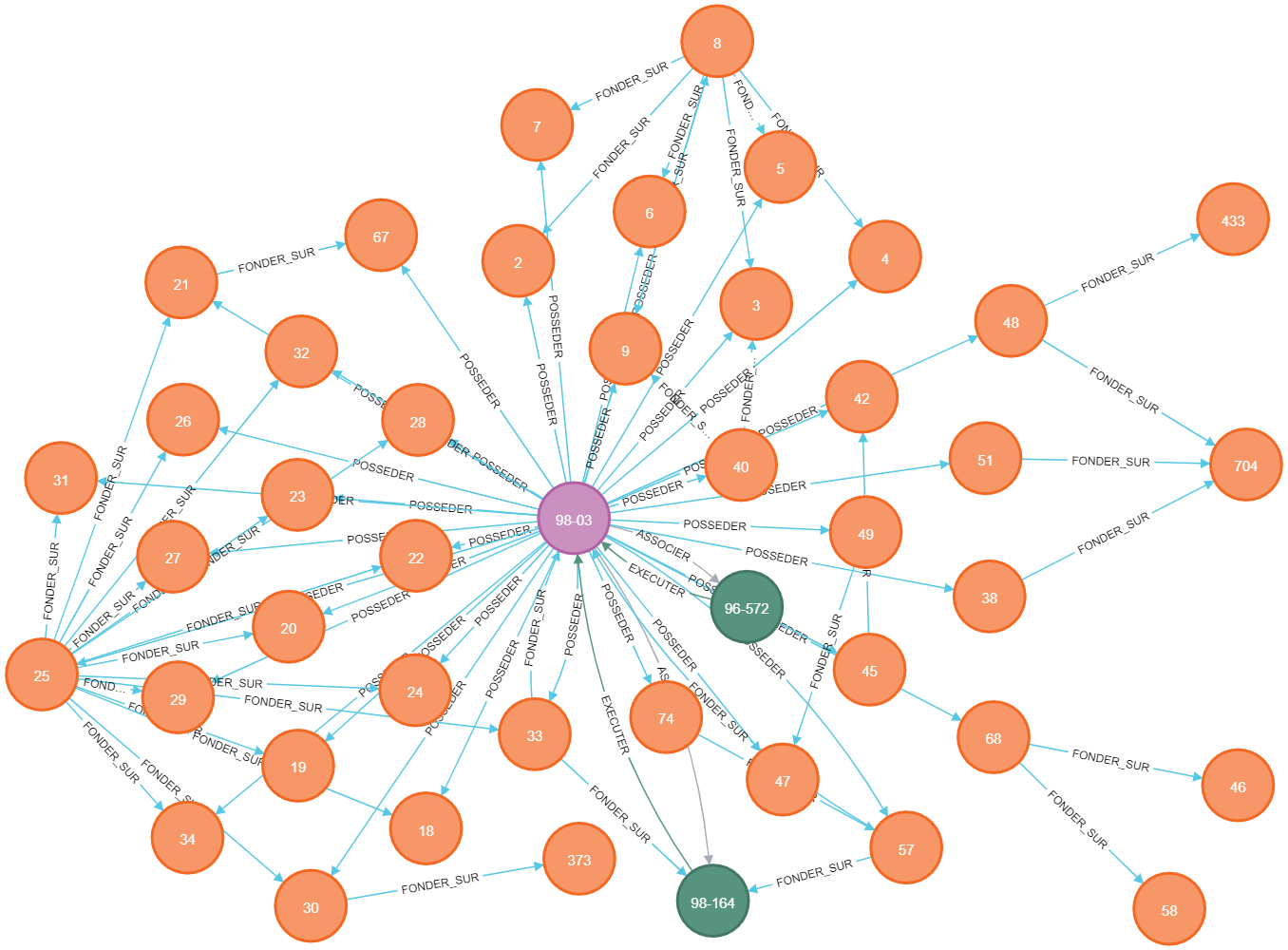}
    \caption{\textbf{Illustration of a Complex Graph}: This diagram depicts the interrelations among decrees (green), Law 98-03 (purple), and its articles (orange). It further elucidates the connections among various articles, extending to three hierarchical levels from Law 98-03's articles.}
    \label{fig:complex_graph}
\end{figure}

\begin{figure}[h]
    \centering
    \includegraphics[width=1\linewidth]{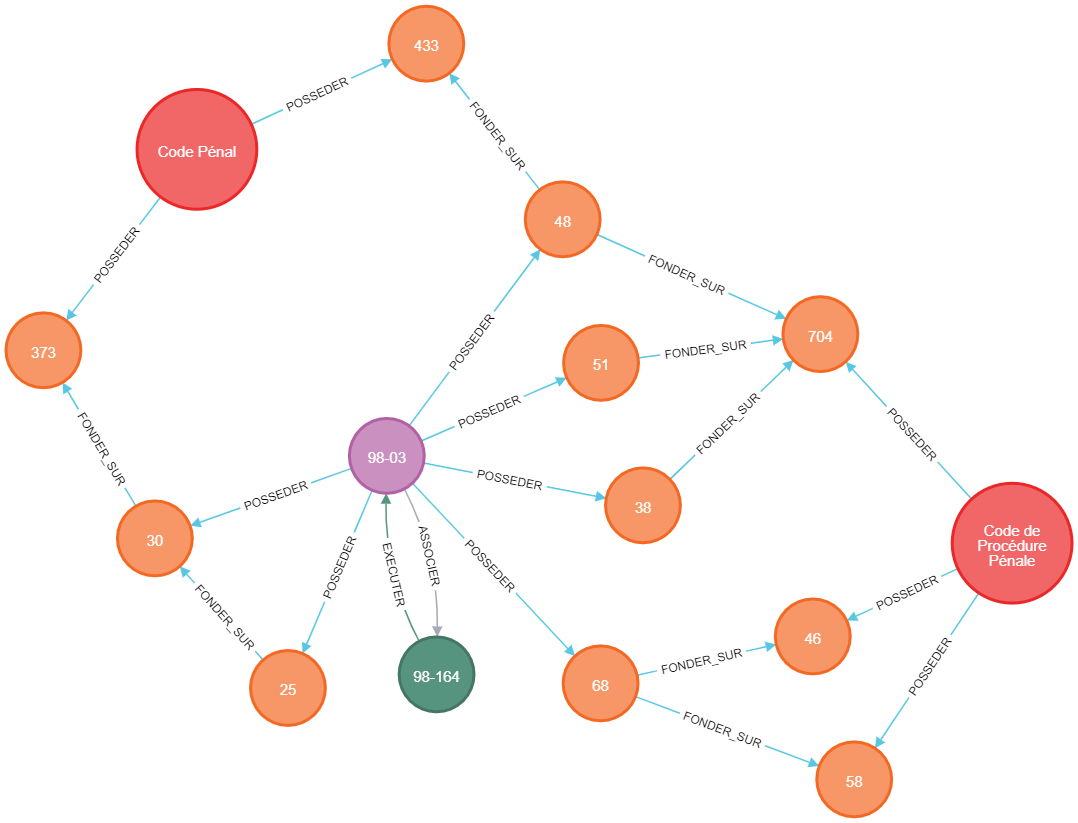}
    \caption{\textbf{Basic Graph Example}: This graph illustrates articles referenced by legal codes (in red) across three hierarchical levels from Law 98-03. 
    }
    \label{fig:simple_graph}
\end{figure}

For experimental purposes, we exemplified graph generation using Cypher. Fig. \ref{fig:complex_graph} depicts a complex graph originating from Law 98-03's articles at three levels deep. The graph becomes simplified when focusing exclusively on articles related to other legal codes, yielding fewer articles and connections, as indicated in Fig. \ref{fig:simple_graph}.

\subsection{Knowledge Triple Extraction}

The experiments were conducted on a Linux 6.1.85+ system utilizing glibc 2.35. The hardware comprised an Intel Xeon 2.00 GHz single-core processor and an NVIDIA Tesla T4 GPU with 16 GB memory. Network performance was assessed using Ookla's Speedtest library directed at a Google Cloud server in Ashburn, VA. Findings indicated a download speed of 9183.89 Mbps, upload speed of 942.81 Mbps, idle latency of 1.91 ms, and 0\% packet loss, confirming robust connectivity for demanding applications like real-time API interactions and extensive data transfers.

Initially, examples were included in the prompt for the extraction of knowledge triples derived from article pairs and their relationships while constructing the graph database. The prompt comprised ten legal articles. We employed the mistral AI model, mistral-large-2411 (Mistral-Large) \cite{jiang_mistral_2023}, to extract references from these articles for contextual enhancement. The extraction process was executed in a single session, requiring merely one minute. Additionally, we identified 24 legal articles for testing, with reference extraction from these taking four minutes, resulting in an average of approximately eight seconds per article.

We assessed the performance of various large language models (LLMs) in extracting knowledge triples. The models evaluated included:

\begin{itemize} \item Mistral AI: mistral-large-2411 (Mistral-Large), pixtral-large-2411 (Pixtral-Large), open-mistral-nemo-2407 (Mistral-Nemo), \item OpenAI: gpt-3.5-turbo-0125 (GPT-3.5-Turbo), GPT-4-0613 (GPT-4), gpt-4o-2024-08-06 (GPT-4o), gpt-4o-mini-2024-07-18 (GPT-4o-Mini) \end{itemize}

All aforementioned models were queried via their respective API endpoints. The resulting knowledge triples were juxtaposed with ground-truth knowledge triples derived from the legal articles in the test, utilizing ROUGE (Recall-Oriented Understudy for Gisting Evaluation) metrics for comparison.

The ROUGE metrics facilitate the automatic evaluation of summary generation models by comparing generated outputs with reference texts to assess similarities. When contrasting generated triples with ground-truth sets, these metrics yield significant insights. ROUGE-1 (R-1) evaluates unigrams for recall based on shared individual words in both generated and ground truth triples. ROUGE-2 (R-2) assesses bigrams to ensure the sequential appearance of words, thereby enhancing the structural coherence of entity relationships. ROUGE-L (R-L) examines the length of the longest common subsequence (LCS) to account for word order effects, permitting matches even when words are non-adjacent. Finally, ROUGE-Lsum (R-Lsum) appraises the overall quality of the triples.

Table \ref{table:rouge_metrics} presents the Estimated Inference Duration (EID) in minutes and seconds, ROUGE metrics for various models, and Number of Parameters in Billions (NPB). Some parameter estimates for OpenAI models are denoted by the symbol $\sim$. Superior EID values are highlighted in red ( \textcolor{score1}{\rule{6pt}{6pt}} ), second best in orange ( \textcolor{score2}{\rule{6pt}{6pt}} ), and third best in blue ( \textcolor{score3}{\rule{6pt}{6pt}} ). The GPT-4o model exhibits the highest ranking in triple generation, followed closely by GPT-4 and Mistral-Large, based on structural and consistency metrics. All three models achieved scores exceeding 80\%, unlike those models that fail to reach a 75\% threshold. Furthermore, GPT-3.5-Turbo, GPT-4o-Mini, and Mistral-Large excel in inference efficiency. Conversely, models with superior triple extraction performance typically exhibit longer execution times, with GPT-4 exceeding 13 minutes and GPT-4o nearly 4 minutes for completion.

\begin{table}[h]
\caption{Comparison between Knowledge Triples Generation Models}
\begin{center}
\resizebox{\linewidth}{!}{%
\begin{tabular}{c@{\hspace{18pt}}c@{\hspace{18pt}}c@{\hspace{18pt}}c@{\hspace{14pt}}c@{\hspace{13pt}}c@{\hspace{15pt}}c}
\hline
\textbf{Mo-} & \multicolumn{6}{c}{\textbf{Metrics}} \\
\cline{2-7} 
\textbf{-dels} & \textbf{\textit{R-1}} & \textbf{\textit{R-2}} & \textbf{\textit{R-L}} & \textbf{\textit{R-Lsum}} & \textbf{\textit{EID}} & \textbf{\textit{NPB}} \\
\hline
Mistral-Nemo & 58.46 & 56.29 & 58.45 & 58.89 & 3m 50s & 12 \\
Mistral-Large & \textcolor{score3}{82.90} & \textcolor{score2}{82.08} & \textcolor{score3}{82.88} & \textcolor{score3}{82.92} & \textcolor{score3}{2m 23s} & 123 \\
Pixtral-Large & 68.25 & 64.85 & 66.90 & 67.44 & 3m 38s & 124 \\
GPT-3.5-Turbo & 73.70 & 71.13 & 73.42 & 73.66 & \textcolor{score1}{1m 18s} & 20 \\
GPT-4o-Mini & 64.10 & 60.11 & 63.44 & 64.07 & \textcolor{score2}{1m 40s} & $\sim$8 \\
GPT-4o & \textcolor{score1}{86.00} & \textcolor{score1}{83.96} & \textcolor{score1}{84.78} & \textcolor{score1}{85.99} & 3m 56s & $\sim$1800 \\
GPT-4 & \textcolor{score2}{83.49} & \textcolor{score3}{82.07} & \textcolor{score2}{83.38} & \textcolor{score2}{83.55} & 13m 29s & $\sim$1800  \\
\hline
\end{tabular}}
\end{center}
\label{table:rouge_metrics}
\end{table}

\section{Discussion}\label{discussion}

\subsection{Extraction of Legal Documents}

In this study, we emphasized the structure of legal documents, noting a lack of access to court judgments and jurisprudence, as evidenced by prior international research \cite{jain_constructing_2022, vuong_constructing_2023}. The challenge of accessing legal documents via the judicial system is compounded by the unavailability of recent updates. This hinders the comprehension of criminal codes and procedures, as tracking changes in laws and decrees is problematic due to insufficient documentation. Thus, there is a pressing need to enhance the accessibility of judicial information.

Moreover, court judgments are secured, necessitating adherence to specific protocols for their appropriate utilization. Presently, access to jurisprudence is limited, with only a recent 2019 collection on land and public domain available \cite{senegal_recueil_2019}. However, many recent documents are only available in scanned formats, complicating data extraction. Consequently, there is a necessity to optimize advanced OCR systems using LLMs or TrOCR \cite{li_trocr_2022} through extensive text datasets in forthcoming research.

Despite these constraints, we successfully gathered complete articles from the majority of the codes. The extracted articles can facilitate the training of advanced machine learning models for precise legal document structure extraction, or be stored in relational databases for reference or similarity analysis via vector databases \cite{reimers_sentence-bert_2019-1, devlin_bert_2019, han_comprehensive_2023-1}.

In addition, the extraction process demonstrates a time complexity of O(n), where n signifies the total number of elements in the DOCX files. This characteristic enables the rapid and efficient processing of Senegalese legal documents, even with large datasets.

\subsection{Graph Database}

Our graph database was systematically constructed through the precise identification of entities within the land and public domain collection via regular expressions. This methodology has yielded a precise and dependable graph database for the collection. Nonetheless, this procedure is labor-intensive and necessitates meticulous recognition of various components to accurately retrieve distinct entities.

Notwithstanding this, as evidenced in this paper, the developed graph database can act as a standard for evaluating future graph databases generated with LLMs or other sophisticated AI methodologies. Furthermore, legal practitioners in Senegal may utilize this graph database for research or in judicial opinions to enhance legal decision-making. Collaborative efforts between legal scholars and data analysis specialists can facilitate ongoing updates within a dynamic legal framework. This initiative aims to create a holistic framework for the automated updating of legal information through a secure information system.

\subsection{Knowledge Triple Extraction}

During our experiments, we conducted a comparative evaluation of various LLMs using ROUGE scores. The GPT-4o model was identified as the most effective in knowledge extraction across all ROUGE metrics. The GPT-4 model followed, yielding relevant results but was outperformed by Mistral-Large in the R-2 metric. Mistral-Large demonstrated superior performance in generating knowledge triples with two consecutive accurate words. It ranked third in R-1, R-L, and R-Lsum evaluations. In summary, GPT-4 outperforms Mistral-Large in triple extraction. These results correlate with model parameters, as higher-performing models generally possess more parameters. Nonetheless, exceptions exist, such as Pixtral-Large, which performs comparably to smaller models despite its 124 billion parameters, due to its image data comprehension during training, unlike Mistral-Large, which focuses solely on textual data.

Conversely, in terms of predicted inference execution time, Mistral-Large is the most efficient model among its peers. It effectively extracts knowledge triples from legal texts with minimal prompting, processing 24 legal articles in just 2 minutes and 23 seconds. However, inference time may not reliably gauge model efficiency, as it is subject to server configurations. Unfortunately, the lack of provided server information by Mistral AI and OpenAI complicates comprehensive timing comparisons. Despite this, estimated inference time can still offer a general assessment of the models' efficiency via the API.

\begin{figure}[h]
    \centering
    \includegraphics[width=\linewidth]{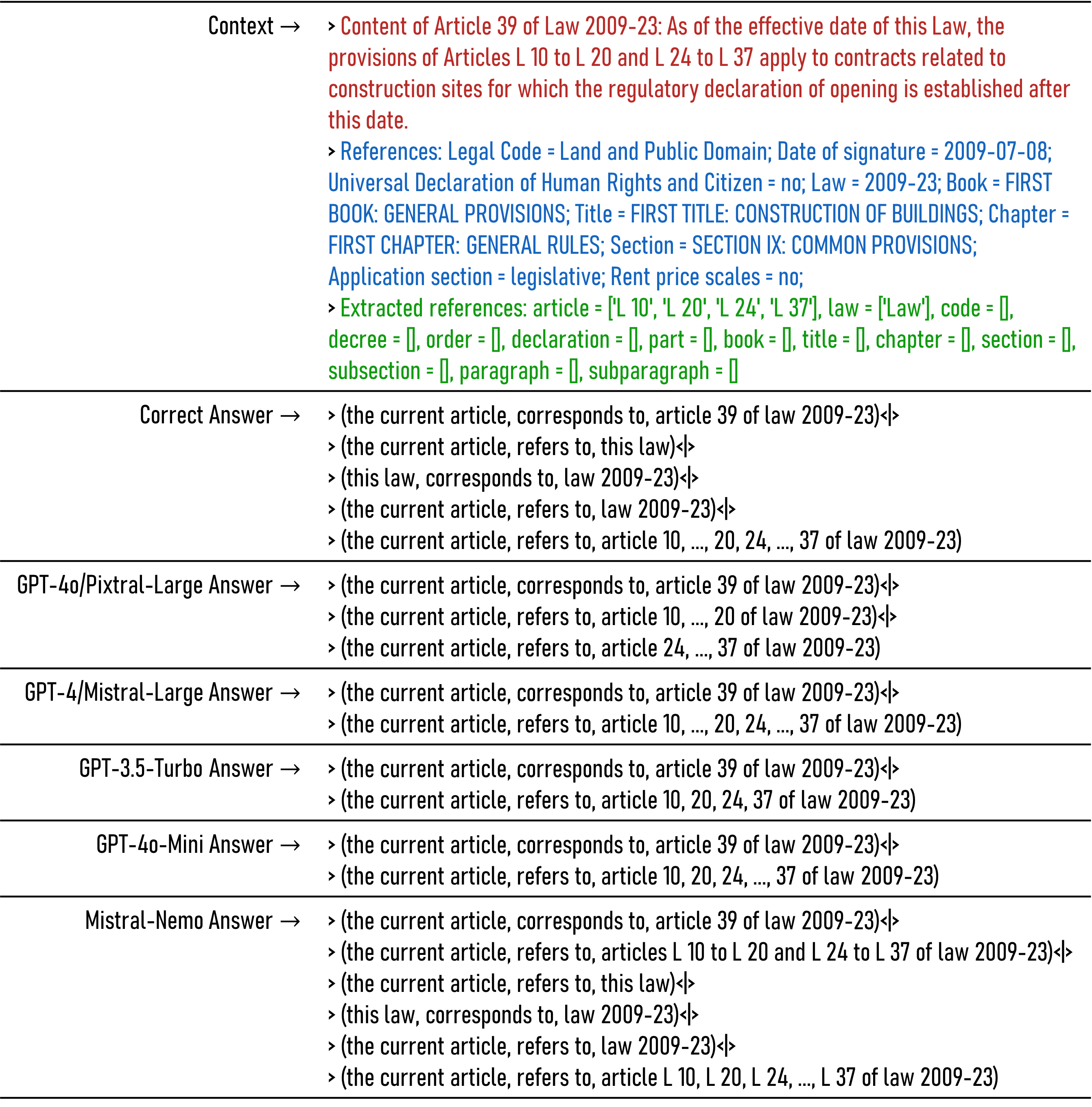}
    \caption{\textbf{Illustration of Answers from Different Large Language Models}: The objective was to analyze the diversity of articles and the primary legislation through deductive reasoning. Only the larger models accurately identified the range, while the Mistral-Nemo model successfully recognized the current law. The symbol "/" serves to distinguish between models that provide identical outputs.}
    \label{fig:example_1}
\end{figure}

To enable a thorough comparison of reasoning capabilities and adherence to triple format across models, the responses for two legal articles are analyzed. The context for generating triples includes the content, metadata, and extracted references of the legal articles. The evaluation of the compatibility of the reasoning and the format is focused on the initial legal article, as shown in Fig. \ref{fig:example_1}.

During evaluation, LLMs must accurately determine the relevant legislation and compile a correct list of cited articles using ellipses (...) as depicted in the Correct Answer. While the GPT-4o and Pixtral-Large models successfully generate diverse article number ranges in structured triples, they fail to clarify the specific essence of "this law".

In contrast, the GPT-4 and Mistral-Large models exhibit proficiency in formatting number ranges within a single triple, indicating their capability in extracting referenced articles through complex representations. However, these models often emphasize articles over other contextual objects.

Other models display varied performance in extracting number ranges. The GPT-3.5-Turbo model omits ellipses for ranges, reflecting a limited grasp of article numbers. Meanwhile, GPT-4o-Mini extracts only one range, revealing constraints in complex triple generation when compared to GPT-4 and Mistral-Large. The Mistral-Nemo model produces two triples for article number ranges, though the formatting is not entirely accurate in the first and incorrectly adds an "L" before individual numbers in the second.

Nonetheless, the Mistral-Nemo model effectively extracts the referenced law, demonstrating a broader focus beyond mere articles. 

\begin{figure}[h]
    \centering
    \includegraphics[width=\linewidth]{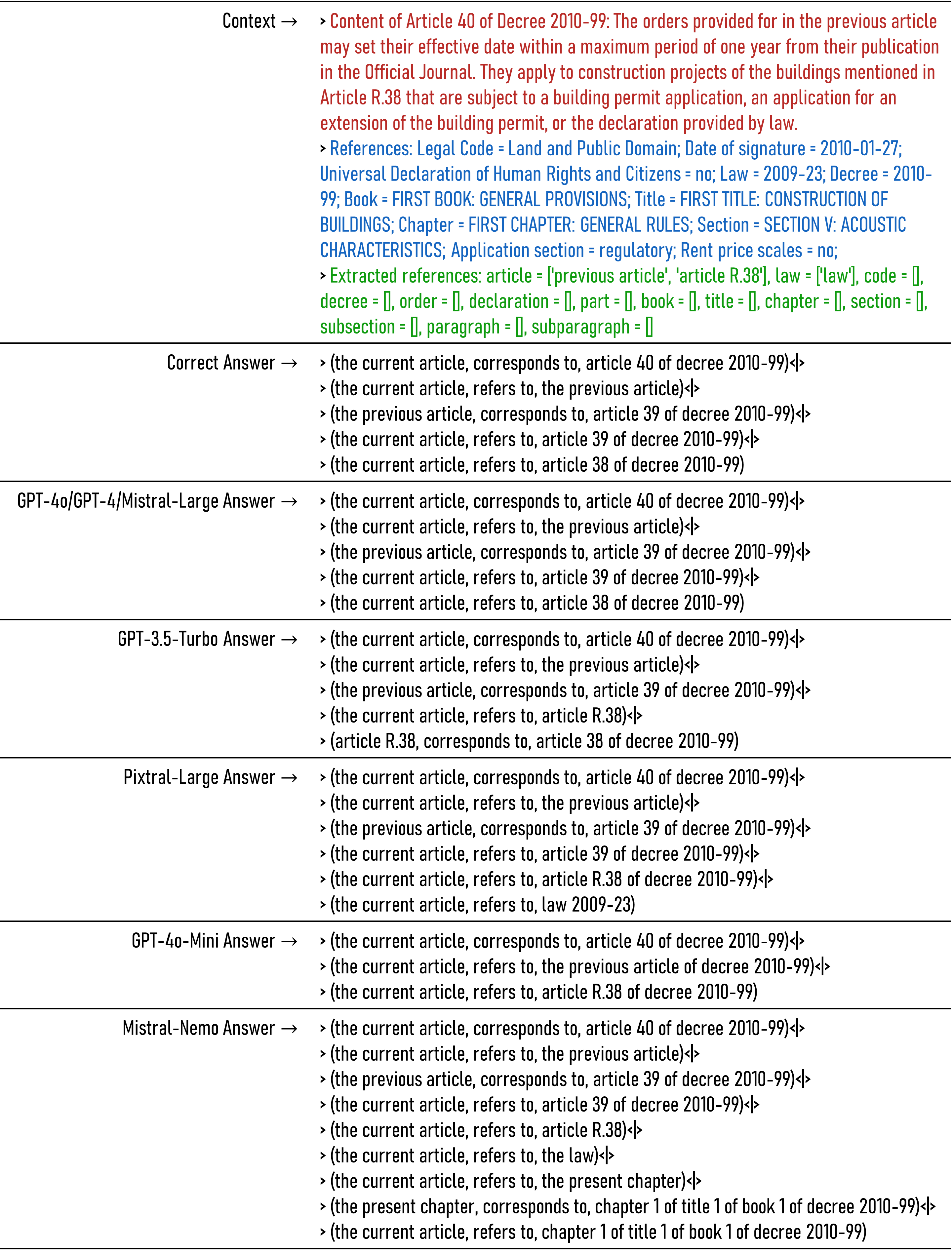}
    \caption{\textbf{Illustration of Answers from Different Large Language Models} This study focused on the retrieval and formatting of a prior article. The expected output was exclusively provided by the GPT-4o, GPT-4, and Mistral-Large models.}
    \label{fig:example_2}
\end{figure}

In Fig. \ref{fig:example_2}, a legal article was analyzed to ascertain the article number of the antecedent article via logical reasoning, while also extracting additional referenced articles with precision.

The GPT-4o, GPT-4, and Mistral-Large models produced congruent results with the Correct Answer, demonstrating strong comprehension of the prompt, robust reasoning capabilities, and compliance with the specified format. Conversely, the GPT-3.5-Turbo model adopted a distinct reasoning approach that deviated from the prompt's guidelines, managing to identify article "R.38" and the "previous article" but failing to complete the final triple with the "refers to" predicate, revealing adequate reasoning but an inability to provide the required triples.

Despite correct extractions, the Pixtral-Large model incorrectly prefixed article number 38 with "R." and extracted an extraneous reference ("law 2009-23") from the metadata, indicating sporadic failures in adhering to the desired format and potential inclusion of irrelevant information.

The GPT-4o-Mini model displayed inadequacies in reasoning and format generation for the legal article, failing to recognize the previous article and mistakenly adding an "R." prefix to article number 38.

Finally, the Mistral-Nemo model successfully identified the previous article but erred in formatting article 38 by including "R." and neglecting the decree number, while also introducing irrelevant elements such as references to "the law" and reasoning concerning the current chapter, which were extraneous to the article's content.

The comprehensive comparison illustrates the variability in model capabilities. Some underperforming models exhibit potential for enhancement through refinement. For instance, GPT-3.5-Turbo, Pixtral-Large, and Mistral-Nemo occasionally generate the desired triples, while more effective models may falter in this regard. This implies that with improved examples and clearer directives, these models could match or exceed the performance of larger models in specific instances. The analysis also indicates that certain models, like GPT-4 or Mistral-Nemo, may produce inaccurate information without adequate contextual input. Notably, GPT-4o-Mini consistently failed to accurately extract correct triples, reflecting inferior knowledge extraction capabilities relative to larger models.



To enhance the reasoning abilities of Large Language Models (LLMs), it is essential to incorporate a broader range of examples in their training. Research should focus on creating detailed prompts with tailored additional examples to improve the generation of knowledge triples. Effective prompt design is crucial for optimal results, while comparative analyses across different LLMs are important for benchmark assessments. This methodology will lay a strong foundation for future projects aimed at developing an advanced legal assistant. This assistant will utilize our LLM-based knowledge triple extraction approach, along with Retrieval-Augmented Generation (RAG) \cite{lewis_retrieval-augmented_2021} and Reasoning and Acting (ReAct) \cite{yao_react_2023} techniques, to provide accurate and contextually relevant responses.

\section{Conclusion}

This research underscores the transformative potential of artificial intelligence (AI) and large language models (LLMs) in Senegal's legal system. Enhancing access to legal documents promotes a more transparent and efficient justice system. The extraction of 7,967 articles and the establishment of a graph database signify substantial advancements in legal information organization.

Our knowledge triple extraction experiments reveal the sophisticated capabilities of LLMs, which are vital for legal practitioners and the public. Using cutting-edge techniques such as Retrieval-Augmented Generation (RAG) and Reasoning and Acting (ReAct) enables the development of an intelligent legal assistant for complex inquiries.

To maintain this framework's viability, collaboration between legal scholars and data scientists is imperative. This initiative enriches the discourse on AI's role in facilitating access to justice, aiming to create an assistant that not only responds to inquiries but also improves the populace's comprehension of legal rights in Senegal.

\bibliographystyle{IEEEtran}
\bibliography{litterature_review_legal_constitution}

\end{document}